\documentclass{article}

\usepackage{arxiv}

\usepackage[utf8]{inputenc} 
\usepackage[T1]{fontenc}    
\usepackage{hyperref}       
\usepackage{url}            
\usepackage{booktabs}       
\usepackage{amsfonts,amsmath,amssymb}       
\usepackage{nicefrac}       
\usepackage{microtype}      
\usepackage{graphicx}

\usepackage{color, colortbl}

\usepackage{mathptmx,amsthm}
\usepackage{xcolor}
\usepackage{multirow}
\usepackage[linesnumbered,ruled,vlined]{algorithm2e}

\newcommand{\ie}{\textit{i.e., }}
\newcommand{\etal}{\textit{et al. }}
\newcommand{\bdesc}{\begin{description}}
\newcommand{\edesc}{\end{description}}
\newcommand{\beqna}{\begin{eqnarray}}
\newcommand{\eeqna}{\end{eqnarray}}
\newcommand{\benum}{\begin{enumerate}}
\newcommand{\eenum}{\end{enumerate}}
\newcommand{\bitem}{\begin{itemize}}
\newcommand{\eitem}{\end{itemize}}
\newcommand{\bfig}{\begin{figure}}
\newcommand{\efig}{\end{figure}}
\newcommand{\cl}{\ensuremath{C_l}}
\newcommand{\cp}{\ensuremath{C_p}}

\newcommand{\gt}{\ensuremath{T_g}}
\newcommand{\st}{\ensuremath{T_s}}
\newcommand{\tv}{\ensuremath{T_v}}
\newcommand{\tvad}{\ensuremath{T_\text{v-ad}}}

\newcommand{\xdownarrow}[1]{%
  {\left\downarrow\vbox to #1{}\right.\kern-\nulldelimiterspace}
}

\newcommand{\ds}[1]{\textbf{DS\textsubscript{#1}}}
\newcommand{\dsi}{\ds{Img}}
\newcommand{\dst}{\ds{Tbl}}
\newcommand{\emd}{\ensuremath{d_\text{EMD}}}
\newcommand{\emdbc}{\ensuremath{\emd{}_\text{-BC}}}
\newcommand{\emdhg}{\ensuremath{\emd{}_\text{-HG}}}
\newcommand{\emdsp}{\ensuremath{\emd{}_\text{-SP}}}

\newcommand{\commenttxt}[1]{}
\newcommand{\mysubheading}[1]{\vspace{0.25cm}\noindent\textbf{#1}\\}
\newcommand{\mysubsubheading}[1]{\vspace{0.25cm}\noindent\textbf{\textit{#1}}}

\definecolor{Gray085}{gray}{0.85}
\usepackage{colortbl}
\newcolumntype{a}{>{\columncolor{Gray085}[.9\tabcolsep][0.9\tabcolsep]}c}

\theoremstyle{definition}
\newtheorem{definition}{Definition}[section]

\SetKwInput{KwInput}{Input}                
\SetKwInput{KwOutput}{Output}              

\SetCommentSty{mycommfont}

\title{Tensor Fields for Data Extraction from Chart Images: Bar Charts and Scatter Plots}
\author{
  Jaya Sreevalsan-Nair\thanks{\texttt{jnair@iiitb.ac.in}} \hspace{1cm} Komal Dadhich \hspace{1cm} Siri Chandana Daggubati\\
  Graphics-Visualization-Computing Lab,\\
  International Institute of Information Technology Bangalore, Karnataka 560100, India \\
  \texttt{http://www.iiitb.ac.in/gvcl} \\\\
  \textit{This is a peer-reviewed article, accepted for publication, prior to its camera-ready version, in}\\
  \textit{"Topological Methods in Visualization: Theory, Software and Applications,"}\\
  \textit{Ingrid Hotz, Talha Bin Masood, Filip Sadlo, and Julien Tierny (Eds.). Springer-Verlag,}
}

\begin{document}
\maketitle

\begin{abstract}
  Charts are an essential part of both graphicacy (graphical literacy), and statistical literacy. As chart understanding has become increasingly relevant in data science, automating chart analysis by processing raster images of the charts has become a significant problem. Automated chart reading involves data extraction and contextual understanding of the data from chart images. In this paper, we perform the first step of determining the computational model of chart images for data extraction for selected chart types, namely, bar charts, and scatter plots. We demonstrate the use of positive semidefinite second-order tensor fields as an effective model. We identify an appropriate tensor field as the model and propose a methodology for the use of its degenerate point extraction for data extraction from chart images. Our results show that tensor voting is effective for data extraction from bar charts and scatter plots, and histograms, as a special case of bar charts.
\end{abstract}


\section{Introduction}\label{sec:introduction}
\begin{figure}[tbp]
\centering
\includegraphics[width=0.9\columnwidth]{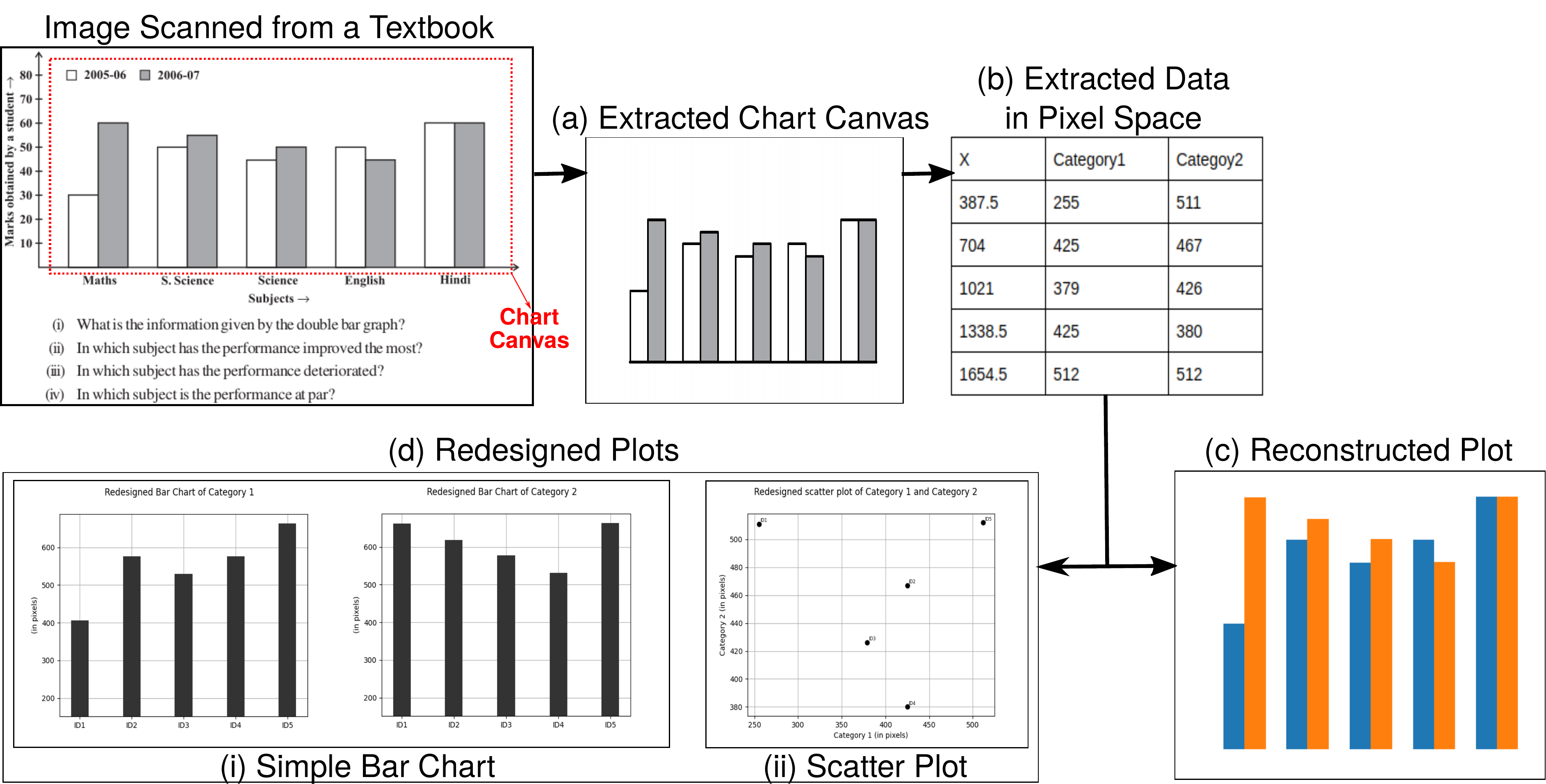}
\caption{Data extraction in pixel space from a chart image using our method, for chart reconstruction and chart redesign. The input image is from a Grade-8 mathematics textbook. (Source: NCERT \protect\url{http://ncert.nic.in/textbook/textbook.htm})}
\label{fig:teaser}
\end{figure}

Charts fall in the intersection of graphicacy (graphical literacy), and statistical literacy. The recent popularity of statistical analysis in data science and ubiquitousness of learning algorithms have made chart graphicacy relevant. Chart comprehension is an outcome of chart graphicacy.  Students often face difficulties in comprehending new chart types when learning new concepts in the visual representation of data, e.g., grouped bar charts ~\cite{burns2009modeling}. Chart images embedded in documents, articles and books often become 
strenuous to analyze further due to unavailability of source data. Each \textit{entity} in a grouped bar corresponds to an ordered proximal placement of a \textit{group} of category-wise bars. Such charts give better communicative signals of (inter-category, intra-entity) trends or relations than those of (intra-category, inter-entity)~\cite{burns2009modeling}. In an example of different academic years (category) grouped together for different subjects (entity) (Figure~\ref{fig:teaser}), the trends in a subject are (inter-category, intra-entity), and those in an academic year are (intra-category, inter-entity). Combinations of intra- and inter-category/entity analysis, e.g., ``what is the difference between the lowest scores in the two academic years?''  require complex interpretation. The inter-category, intra-category, and combination questions are best communicated by grouped bar charts, a set of simple bar charts, and a scatter plot, respectively (Figure~\ref{fig:teaser}). Such chart redesign is possible only with the data extracted from the original charts, that are usually available in raster image format scanned from subject textbooks, e.g., mathematics, science, economics, etc. Thus, we are interested in automatic data extraction from the chart images.

In 1999, Kimura, a Japanese educator, had proposed a six-level scheme for statistical ability using graphs (charts)~\cite{aoyama2003graph}. Even though traditionally ``graph'' and ``chart'' are interchangeably used, we use the latter to disambiguate from the graph data structures used in computer science. The scheme progresses from a basic Level-A with four sub-levels (A1-A4) to the advanced Level-F, where the knowledge from a chart combined with other relevant information is used for new inferences. The sub-levels A1 refers to the basic reading of chart image by reading title, unit and values which further is enhanced with sub-level A2 by enabling the students to read key features, e.g., minimum and maximum values and value differences etc. The sub-level A3 is for comparing information from two different charts, and A4 is in reading trends in charts. Levels B-F involves knowledge outside of the chart itself to understand the chart, e.g. data sources, complex statistical computations, current affairs, etc. The computational model encompassing the six levels is needed for completely automated chart interpretation that can address a pertinent need of assistive solutions for chart graphicacy for the differently-abled, including the visually impaired~\cite{burns2009modeling,choi2019visualizing}. 

Appropriate computational models complement the machine interpretation of charts~\cite{poco2017reverse} and learning algorithms for chart interpretation. The focus of the models discussed here, are on understanding charts using WYSIWYG (what you see is what you get) features from its images. In general, the challenges for developing such models for chart images lie in the vastness of the charts' design space, attributed to the different types, formatting, applications, and usability patterns. Despite the recent advances in the use of deep learning for chart interpretation~\cite{choi2019visualizing,cliche2017scatteract}, these solutions do not still cover the vast design space of charts~\cite{liu2013review}. The use of machine learning in chart interpretation has been limited in its applications owing to the inadequacy of a single model. The learning models generated using procedurally generated datasets do not account for artefacts present in readily available images (e.g., those available on the internet)~\cite{cliche2017scatteract}. Reuse of object detection models, e.g. YOLO that have been developed for all image types leads to detection errors in chart images, accounting for about 21.6\% cases of data extraction failure~\cite{choi2019visualizing}. A generalized  computational model for chart image processing that enables feature extraction can ideally complement existing learning models.

\begin{figure}[tbp]
\centering
\includegraphics[width=0.9\columnwidth]{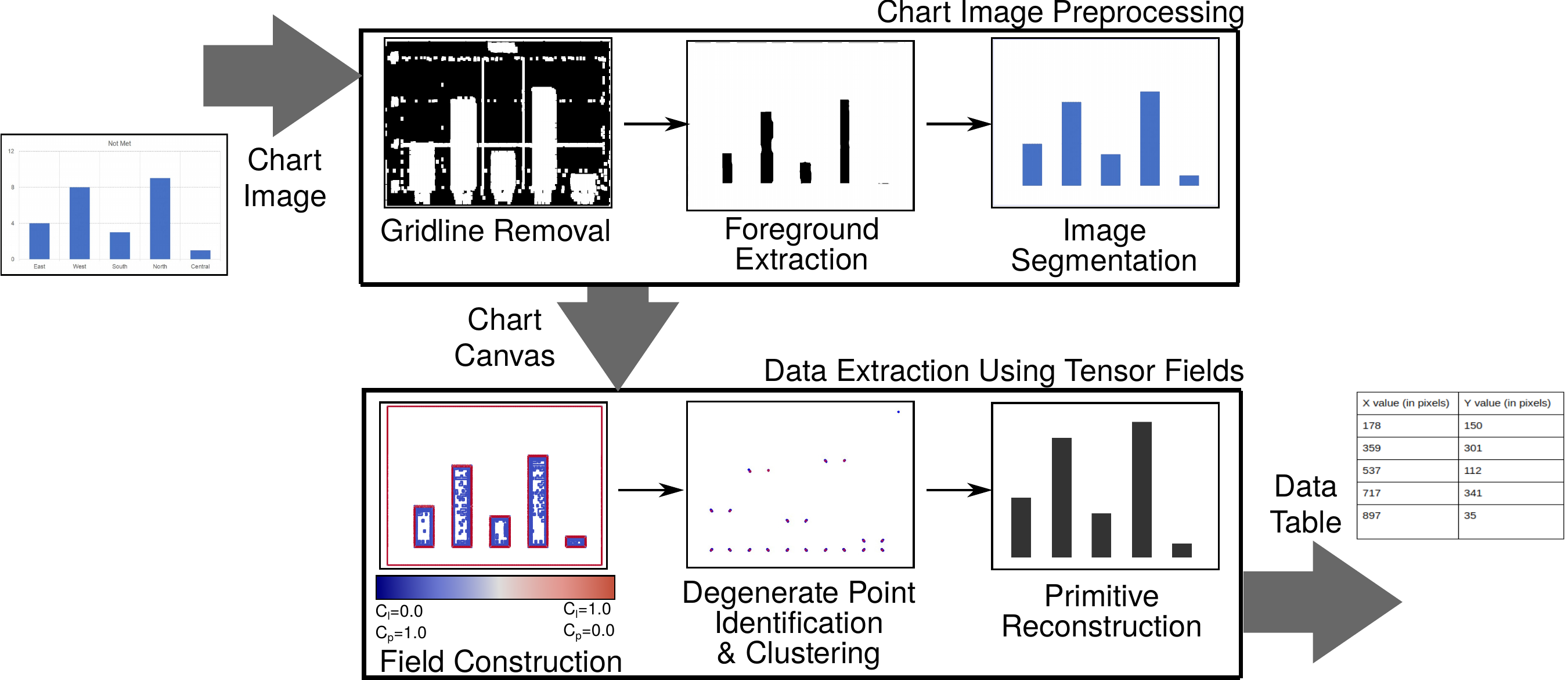}
\caption{Our proposed workflow of data extraction from a chart image using positive semidefinite second-order tensor field of local geometric descriptors. (Image source: \protect\url{https://excelnotes.com/how-to-add-gridlines-to-a-chart/})}
\label{fig:blockdiagram}
\end{figure}

In the cognitive science of chart interpretation, the global precedence principle states that local properties (color, geometry) of a visual object are processed only after its global properties~\cite{wagemans2012century}. However, computationally building global properties from local ones is more tractable~\cite{guy1993inferring} than vice-versa. Structure tensor and tensor voting are both effective for pixel-wise local structure estimation in images~\cite{moreno2012adaptation}. Tensor voting has also been used to generate a positive semidefinite second-order tensor field in 3D point clouds for local geometric descriptors, that used for semantic classification~\cite{sreevalsannair2017local}. Similarly, we propose to use these tensor fields for local structure estimation in charts for data extraction (Figure~\ref{fig:blockdiagram}), using its degenerate points. We focus on two different chart types, namely, simple bar charts and scatter plots, that have a bijective mapping between data items and  geometric objects (\ie bars and scatter points, respectively). But they differ in their source data distributions, namely, univariate and bivariate distributions, respectively. Similar work has so far been dealt with either a single chart types~\cite{cliche2017scatteract,siegel2016figureseer} or multiple types of a univariate distribution~\cite{choi2019visualizing,poco2017reverse}. Thus, our contributions are in:
\bitem
\item using a positive semidefinite second-order tensor field as a computational model for processing chart images, with a focus on bar charts and scatter plots, 
\item identifying an appropriate tensor field representing the geometric information from the images for data extraction,
\item using degenerate points of the tensor field for data extraction.
\eitem
Overall, our contributions result in a computational model to automate levels A1 and A2 in Kimura's scheme, from chart images, as they are the only levels pertaining to the geometric information.

\section{Related Work}\label{sec:litsurvey}
We discuss the relevant state-in-the-art in characterizing chart interpretation and automated systems for chart analysis.

\mysubsubheading{Chart Interpretation and Analysis:} While there have been several studies done in chart interpretation in parts, or whole for few chart types, there is still a gap in standardized and generalized methods for chart interpretation, which will work for more than one type of chart~\cite{liu2013review}. Liu \etal concluded that prior human knowledge is an important element, which makes a case for machine learning algorithms. Our work will be useful in generating effective feature vectors for supervised learning models, which perform chart classification, chart segmentation, and chart segment classification, etc., ~\cite{choi2019visualizing}. Chart interpretation has been researched extensively in cognitive science~\cite{hegarty2011cognitive} and document engineering~\cite{huang2007system}.  The findings in the latter continue to be extensively used in separating text and graphics in charts using Optical Character Recognition (OCR) techniques~\cite{choi2019visualizing,poco2017reverse}.

Our proposed method has been developed based on the information gathered from the vast literature in cognitive science in the context of chart analysis. The use of proximity of displayed variables validates our use of spatial locality-based approaches for an integral display, specifically the scatter plot~\cite{jones1990display}. Wickens and colleagues have extensively studied separable and integral displays~\cite{jones1990display} and considered bar charts as separable displays. Separable displays are known generally to be not effective for information integration.  However, bar chart analysis was found to be useful for information integration~\cite{wickens1995proximity}, e.g., when using the proximity of center points of top-line of bars to visualize trends. Bar charts and other charts have been found to demonstrate properties of integral, configural, and separable/object displays depending on the mapping of the variable~\cite{bennett1992graphical}. Unlike the use of spatial proximity in chart analysis, we use spatial proximity differently in bar charts to identify ``corners'' of the top line of bars and scatter plots to locate the scatter points. The spatial locations are crucial for data extraction, thus substantiating the use of spatial locality.

\mysubsubheading{Systems for Chart Analysis:} There have been several systems recently developed for various aspects of chart interpretation: text localization and recognition~\cite{poco2017reverse}. A more comprehensive system for the visually impaired takes in chart images in a webpage through a browser extension and outputs a data table and accessible interactive charts~\cite{choi2019visualizing}. Choi \etal perform data extraction differently for bar charts, pie charts, and line charts. For bar charts,  a combination of a Darknet neural network outcomes for object detection and an OCR system for text extraction is used. For pie charts, a combination of color to item matching using the OCR system and finding the proportion of pixels corresponding to a particular color within a bounding box of the circle is used. For line charts, pixels with the same color and across spans between consecutive tick values are used to identify y-values at each x-tick mark. However, since the geometric information captured in these different methods is extracted from image data, e.g., color, we hypothesize that a geometry-aware method can be generically used for identifying key pixels in the chart images.
FigureSeer~\cite{siegel2016figureseer} is another end-to-end framework for summarization of line charts. It is implemented using a convolutional neural network that compares image patches. For data extraction, an optimal solution for the path-finding problem is used. Beagle~\cite{battle2018beagle} is a system for classifying charts found as visualizations on the internet. ReVision~\cite{savva2011revision}, Scatteract~\cite{cliche2017scatteract}, and a bar chart-based method~\cite{al2015automatic} are examples of automated systems for data extraction from selected chart type(s).

\mysubsubheading{Feature Extraction:} There has been an even lesser focus on generating feature vectors for chart analysis. In ReVision~\cite{savva2011revision}, a feature vector is created using a clustering method for determining codebook patches in an image and finding similar patches to determine a histogram of activated patches. The feature vector has been further used for chart classification. The data extraction has been done using pixel-based methods, driven by the knowledge of chart layouts for bar charts and pie charts. Our proposed tensor field-based method is generically used across two different chart types to identify critical points to localize ``objects'' for value extraction. These critical points can serve as features.

Overall, chart analysis is a problem solved in pieces, and our work is novel in geometric analysis of chart canvas for data extraction. Our work would be the closest to the use of Hough transform in image space to recognize bar charts~\cite{zhou2000hough}, in the context of deriving features in image space and finding generic patterns.

\section{Background on Local Geometric Descriptors}\label{sec:bg}
Local geometric descriptors, which encode the local geometry of each entity in a dataset, can be represented in the form of positive semidefinite second-order tensor fields.
Descriptors such as structure tensor~\cite{knutsson1989representing} have been used for corner detection, shape analysis, and feature tracking, and tensor voting~\cite{medioni2000tensor} has been used in images widely for segmentation~\cite{jia2004inference}. 

\mysubsubheading{Structure Tensor:} Structure tensor $\st$ encodes the directionality of the gradient of the local neighborhood. $\st$ is computed from the gradient tensor, $\gt=G^TG$, using the gradient vector, $G=\begin{bmatrix} \frac{\partial I}{\partial x} & \frac{\partial I}{\partial y}\end{bmatrix}$, at a pixel with intensity $I$.

Difference kernels, such as Sobel operator, are used for determining discretized differences required for computing $G$ in images. Applying Gaussian convolution to $\gt$ gives $\st$. For a Gaussian function with zero mean and standard deviation $\rho$, $\st=G_\rho*\gt$, where $*$ is a 2D convolution operator, in the case of 2D images.

\mysubsubheading{Tensor Voting:} Tensor voting is a technique of determining global perceptual information organization by garnering votes at each entity based on the normal tensors of its local neighbors~\cite{medioni2000tensor}. It is especially effective for applications requiring a global context, such as image segmentation. Here, the votes are aggregated component-wise, namely, stick-, plate-, and ball-tensors, in 3D data, and stick- and ball-tensors in 2D. Thus, tensor voting gives a positive semidefinite aggregated tensor $\tv$ of propagated votes, that are positive semidefinite normal tensors. 

Gradient information can be used to compute tensor voting by using $\gt$ to initialize the stick-tensor component specifically~\cite{moreno2012adaptation}. For grey-scale images, the $\gt$ is computed for any one of the RGB channels; and for color images, the $\gt$ is computed separately for the three channels, and the tensor votes across all neighbors and all channels are aggregated by summation. Additionally, the tensor voting in natural images has been approximated to stick-tensor votes alone, since the percentage of pixels with a ratio of eigenvalues $\big(\frac{\lambda_0}{\lambda_1}>0.1\big)$, where eigenvalues of $\tv$ at each pixel are, such that, $\lambda_0\ge\lambda_1$, has been found to be considerably low ($\sim$10\%)~\cite{moreno2012adaptation}. Recently, the closed-form analytical solution of tensor voting has been determined~\cite{wu2016closed}\footnote{This is the version of the paper with clarification of comments given on a previous publication~\cite{wu2011closed}.}. The tensor vote cast at $x_i$ by $x_j$ using a second-order tensor $K_j$ in $d$-dimensional space is:
\centerline{$S_{ij}=c_{ij}R_{ij}K_jR_{ij}'$, where $R_{ij}=(I_d - 2r_{ij}r_{ij}^T)$, and  $R_{ij}'=(I_d-\frac{1}{2}r_{ij}r_{ij}^T)R_{ij}$.}\\
$I_d$ is the d-dimensional identity matrix; direction vector $r_{ij}=\hat{d_{ij}}$, where the distance vector $d_{ij}=x_j-x_i$; and $c_{ij}=\text{exp}\big(-\big(\sigma_d^{-1}.\|d_{ij}\|_2^2\big)\big)$, where $\sigma_d$ is the scale parameter. The gradient tensor $\gt$ can be used as $K_j$~\cite{moreno2012adaptation}. In the closed form, if a generic tensor field is used as $K_j$, then a voting field is not required~\cite{wu2016closed}. Thus, tensor voting is no longer separately computed as the different components of a stick-, plate-, and ball-voting fields, and aggregated. Once $S_{ij}$ is computed for a point with each of its neighbors in a (von Neumann or 4-) neighborhood $\mathbb{N}_4$. Then the votes are aggregated across all the neighbors using summation. Using $\sigma_d=4$, based on neighborhood size, we get the aggregated positive semidefinite second-order tensor as:\\
\centerline{$\tv=\sum\limits_{k=0}^{(d-1)}\sum\limits_{j\in\mathbb{N}_4} S_{ij}(d)$.}

\mysubsubheading{Anisotropic Diffusion:} Since $\tv$ propagates the normal tensor votes, the aggregated tensor is in the normal space. However, we need tensors that encode geometry of objects (such as bars and points) in the image, implying the tensors must be tangential to the object boundaries. Thus, the tensor $\tv$ must be transformed into tangent space, which can be done using anisotropic diffusion~\cite{sreevalsannair2017local,wang2013anisotropic}. Adapting to the 2D case, using eigenvalues of $\tv$, $\lambda_0 \ge \lambda_1$, and corresponding eigenvectors $v_{0}$ and $v_{1}$, respectively, the tensor after anisotropic diffusion using diffusion parameter $\delta$, is: \\\centerline{$\tvad=\sum\limits_{k=0}^1 \lambda_k'.v_{k}v_{k}^T$, where $\lambda_k'=\text{exp}\Big(-\frac{\lambda_k}{\delta}\Big)$.}
  Diffusion parameter ($\delta=0.16$) is a widely used parameter setting~\cite{ sreevalsannair2017local,wang2013anisotropic}.

  \mysubsubheading{Saliency Map Computation:} The eigenvalue decomposition of the local geometric descriptor of a 3D point determines if the point is part of either a line-, surface, or point-type feature~\cite{keller2011extracting, kumari2015interactive}. Similarly, in 2D image space, the eigenvalue decomposition of the local structure estimator, such as $\tvad$, determines the probabilistic geometric classification of pixels to the line- and point-type features. Hereafter, we use ``local structure estimator'' and ``local geometric descriptor'' interchangeably. The saliency maps~\cite{medioni2000tensor} summarize the likelihood of the point or pixel belonging to these classes at the end of the tensor voting process.
  
  Thus the likelihood of a pixel being a point-type feature, $\cp$, can be derived from the junction-map and that of a line-type feature, $\cl$, from the curve-map. It is a probabilistic classification, and hence, $\cl+\cp=1.0$.

  Adapting the computation of anisotropy in superquadric tensor glyphs~\cite{kindlmann2004superquadric} has been found suitable for saliency maps of 3D points~\cite{sreevalsannair2017local}. Similarly, in 2D images, we get the saliency maps at each pixel, namely curve-map $\cl$ and junction-map $\cp$, as: \\
  \centerline{$\cl=\frac{\lambda_0-\lambda_1}{\lambda_0+\lambda_1}$ and $\cp=\frac{2\lambda_1}{\lambda_0+\lambda_1}$.}\\
  where we have eigenvalues of $\tvad$ of the pixel with $\lambda_0\ge \lambda_1$. Thus, bringing together tensor field topology and probabilistic geometric classification, a pixel with $\cp\approx 1.0$ is a degenerate point, attributed to the anisotropic local neighborhood.

\section{Our Proposed Method}\label{sec:method}
The aim of our work is twofold -- firstly, we propose the use of local structure estimation to generate a computational model for chart interpretation, and secondly, we show the effectiveness of this model for data extraction from simple bar charts and scatter plots. We run experiments on a variety of charts to study the effectiveness of our model. For this work, we focus exclusively on data extraction purely using geometry, without extracting contextual information from text, axes, and legend. Hence, we work with only with the relevant region in the chart image, which warrants defining chart image components. The local geometric descriptors of the pixels in this region are the tensor field, and it is used as a computational model for chart images. We explain the observed behavior of the proposed tensor model for geometric entities, namely bars and scatter points, and its degenerate points. The understanding of the field is important for deciding the data extraction workflow (Figure~\ref{fig:blockdiagram}) and performing the error analysis of the model. This section has three main parts on chart image components, our proposed computational model, and the data extraction workflow.

\mysubheading{A. Chart Image Components:}
Here, we connect the characteristics of chart images with tensor field topology. Hence, we first define the terminology pertaining to the charts and descriptions of relevant parts of the charts. As mentioned in Section~\ref{sec:introduction}, the digitized image format of a chart is called the \textit{chart image}.

\begin{definition}{} The \textit{chart data} is the data used for plotting in a chart. The chart data is a uni- and bi-variate distribution in simple bar charts and scatter plots, respectively. 
\end{definition}
\begin{definition}{} A two-dimensional geometric primitive or mark (in information visualization parlance) in the chart that encodes chart data is defined as a \textit{chart object}. Bars and scatter points are chart objects in bar charts and scatter plots, respectively.
\end{definition}
\begin{definition}{} Two pixels $p$ and $q$ are said to be \textit{connected} if $q\in N_4(p)$ or $a\in N_8(p)$, in 4-neighborhood ($N_4$) or 8-neighborhood ($N_8$) of $p$. A set of connected pixels in an image is defined as a \textit{connected component}. A connected component in a chart image is called a \textit{chart-image-component}.
\end{definition}
\begin{definition}{} The \textit{chart canvas} is the rectangular region of the chart image corresponding to the plot, which is the graphical representation of chart data.
\end{definition}
\begin{definition}{} A chart-image-component that corresponds to a chart object and is a subset of chart canvas is defined as a \textit{chart-object-component}, respectively. Connected components of more than one chart object components are called \textit{chart-object-clusters}. For sake of simplicity, we refer to isolated chart-object-components also as chart-object-clusters.
\end{definition}
\begin{definition} The {\textit{component boundary}} of a chart-object-cluster is a chart-image-component whose pixels belong to the chart-object-cluster, and one or more their $N_8$ neighbors do not belong to the chart-object-cluster. The \textit{borders} applied to individual chart objects using the plotting tool are subsets of the component boundary. 
\end{definition}
\begin{definition} The {\textit{component interior}} of a chart-object-cluster is the chart-image-component whose pixels and their entire $N_8$ neighborhoods belong to the chart-object-cluster. The union set of component boundary and component interior of all chart-object-cluster in chart canvas becomes the \textit{foreground} of the chart canvas.
\end{definition}
\begin{definition} The {\textit{component exterior}} of a chart-object-cluster is the difference set of the chart canvas and the union of the component boundary and the component interior of the chart-object-cluster. The intersection set of component exteriors of all chart-object-clusters in chart canvas is the \textit{background} of the chart canvas. 
\end{definition}

\mysubheading{B. Computational Model:}
The size (height) and position of the chart objects encode the information from chart data in bar charts and scatter plots, respectively. Methods for detection and localization of chart-image-components rely on object detection methods used in image processing~\cite{choi2019visualizing}. For objects found in any raster image, its boundary and position can be extracted using the local descriptors of pixels using edges and junctions~\cite{kothe2003integrated}. In our case, we are interested in identifying and localizing chart-object-clusters in bar charts and scatter plots for data extraction. We propose identifying component boundaries for the chosen chart types. In the chart canvas, the gradient change along the component boundaries is captured effectively using the gradient tensor $\gt$. Thus, we propose to use local geometric descriptors derived from $\gt$  as options for tensor fields to model the chart canvas. The descriptors are the structure tensor, $\st$, and those from the tensor voting field after anisotropic diffusion, $\tvad$. The use of positive semidefinite second-order tensor fields formalizes a generalized model across different chart types, with the potential scope of global-local feature extraction. 

The component boundary of chart-object-clusters manifests as line-type features, \ie with high $\cl$ of the local geometric descriptors. However, the pixelated boundaries also lead to occurrences of point-type features, where $\cp$ is high. The point-type features are synonymous with the degenerate points in tensor fields, where the eigenvalues of the tensor are equal to each other~\cite{delmarcelle1994topology}. In local geometric descriptors of 3D point clouds, these have been loosely referred to as point-type features~\cite{sreevalsannair2017local}, or as critical points~\cite{keller2011extracting}. The junction points in images are known to have high ball saliency~\cite{mordohai2004junction} or high junction saliency~\cite{guy1997inference}. The junctions are also synonymous with degenerate points in tensor fields, as they are the intersection of multiple edges and, thus, have anisotropy in its local neighborhood. Overall, the degenerate points are relevant here for extracting the component boundary in both the two chart types.
\begin{itemize}
\item In bar charts, the corners of the bars are captured using junction maps or degenerate points. With the knowledge of the orientation of the bars (vertical/horizontal), it is straightforward to determine the distance between degenerate points in the corners along the appropriate axis, which gives the size of the bar in pixel space. 
\item In scatter plots, the centroid of the degenerate points in the component boundary of the chart-object-clusters gives the location of the chart object, \ie the scatter point, in pixel space. The centroid of degenerate points in the boundary can be treated equivalent to the centroid of the component interior of the scatter points.
\end{itemize} 
Thus, the degenerate points of the proposed tensor field encode the information in the chart data of our chosen chart types. We label the pixels with $\cp>\tau_{cp}$ for a threshold $\tau_{cp}$ as degenerate points. For our experiments, we use $\tau_{cp}=0.6$.

\mysubheading{C. Data Extraction Workflow:}
We first look at data extraction from the tensor fields, and then we look at the requirement for chart image preprocessing for improving the tensor fields themselves, as shown in the blocks in Figure~\ref{fig:blockdiagram}.

\mysubsubheading{Data Extraction Using Tensor Fields:}
In the tensor field of the locality of the corners of bars or the component boundary of scatter points, the degenerate points tend to form local clusters, some of them tend to be weak. Since our requirement for extracting specific locations of geometric primitives, we perform two-step postprocessing of degenerate points. Firstly, We discard weak degenerate points by thresholding based on a tensor invariant, namely the trace. Here, we consider the degenerate points with unity-based normalized trace $T<\tau_{wd}$, for a threshold $\tau_{wd}$, as weak. In our experiments, we have used $\tau_{wd}=0.005$ for bar charts, and $\tau_{wd}=0.01$ for scatter plots. Secondly, we use density-based clustering to consolidate the degenerate points in the corners of the bar as well as to separate individual scatter points. As most clustering algorithms depend on the hyperparameters based on cluster output, e.g., the number and shape of clusters, we require a method that can work without the prior knowledge of such variables. Hence, we choose DBSCAN (Density-Based Spatial Clustering of Applications with Noise)~\cite{ester1996density} for clustering degenerate points. The DBSCAN algorithm is based on this intuitive notion of ``clusters'' and ``noise,''  where each point in a cluster must have at least a minimum number of local neighbors. DBSCAN output is influenced by the distance between clusters ($\epsilon$) and the minimum number of points in a cluster ($n$). In our method, we have found that these parameters of DBSCAN need to be evaluated for each image individually owing to the variability present in the chart images. We use our visualization of degenerate points to decide these values for each image. 

For visualizing the tensor field of the chart canvas, we use ellipsoid glyphs colored based on saliency maps. We also use a dot plot to visualize the saliency maps at all pixels in the chart canvas image and superimpose degenerate points on the original chart image to identify DBSCAN parameters. We use the colorblind safe divergent color palette, namely the coolwarm palette, for $\cl$ values from 0 to 1. This color map also reveals $\cp$ values, as $\cl+\cp=1.0$, thus, identifying the blue pixels as degenerate points. We also use the visualization to evaluate the performance of $\st$ and $\tvad$ tensor fields as an appropriate computational model.

Our workflow (Algorithm~\ref{alg:workflow}) for data extraction from tensor fields includes tensor field computation in a chart canvas and chart-type-dependent data extraction. Since we do not have the contextual information of the plots, the data is extracted in pixel space. We use the set of centroids of degenerate point clusters $D_{cq}$ for data extraction. For bar charts, we sort the centroids of first by x, and then by y. We use a scanline algorithm in increasing value of x to determine missing points based on the pattern of occurrence of corners by repeating x- and y-values. We thus use a \textit{rule-based occurrence pattern} to find missing points, e.g., simple bar charts, and histograms have distinct patterns. Finding the range of y-intervals at each unique x value gives the univariate distribution of the chart data. On the other hand, for scatter plots, the centroids of points in $D_{cq}$ provide the bivariate distribution of the chart data. Since we do not have a mechanism for filling missing values in scatter plots in our current work, we encounter omission errors (type-2 errors or false negatives), unlike in the case of bar charts.

\begin{algorithm}[htbp]
\caption{Data extraction using tensor fields from chart images}
\label{alg:workflow}
\DontPrintSemicolon

\KwInput{Chart image $C_i$, chart-type $C_t$}
\KwOutput{Data table $D$}
Initialize $D$ $\leftarrow$ $\varnothing$ \;
Initialize $S_\textrm{deg-pt}$ $\leftarrow$ $\varnothing$\tcp*{Set of degenerate points} \;
Initialize $D_{cq}$ $\leftarrow$ $\varnothing$ \tcp*{Cluster centroids of degenerate points}\;
$C_c$ $\leftarrow$ chart-canvas-extraction($C_i$) \tcp*{from Algorithm 1}\;
\For {pixel i in $C_i$}
{
  $N$ $\leftarrow$ find-$N_8$-local-neighborhood(i)\;
  $T_\textrm{geom}$ $\leftarrow$ compute-tensor(descriptor-type, $N$) \tcp*{Local geometric descriptor}\;
   $C_l, C_p$ $\leftarrow$ compute-saliency-map($T_\textrm{geom}$)\;
   \tcc{Check if the pixel is a strong degenerate point}\;
    \If {$C_p$$>$$\tau_{cp}$ and trace$(T_\textrm{geom})$$>$$\tau_{wd}$}          
    {
         $S_\textrm{deg-pt}$ $\leftarrow$ set-union($S_\textrm{deg-pt}$, i) \;
     }
}
$C_\textrm{deg-pt}$ $\leftarrow$ DBScan($S_\textrm{deg-pt}$) \tcp*{Cluster degenerate points}\;

\For {cluster q in $C_\textrm{deg-pt}$}
{
     $C_q$ $\leftarrow$ compute-centroid(q) \;
     $D_{cq}$ $\leftarrow$ set-union($D_{cq}$, $C_q$)\;
}

\If {$C_t$ is bar-chart}
{
   $D_{cq}$ $\leftarrow$ set-union($D_{cq}$, missing-points) \tcp*{Rule-based occurrence patterns}\;
  $D_{cq}$ $\leftarrow$ sort-first-by-x-and-sort-second-by-y($D_{cq}$)\;
  \For {unique x-value in (x,y) in $D_{cq}$}
  {
        $\delta_y$ $\leftarrow$ find-y-intervals(x, $D_{cq}$)\;
        $D$ $\leftarrow$ set-union($D$, (x, $\delta_y$)) \tcp*{Add to data table}\;
  }
}
\ElseIf {$C_t$ is a scatter-plot}
{
   \For {(x,y) in $D_{cq}$}
   {
        $D$ $\leftarrow$ set-union($D$, (x,y)) \tcp*{Add to data table}\;
   }
}

\KwRet $D$
\end{algorithm}

\mysubsubheading{Chart Image Preprocessing:}
We need to preprocess the chart images before computing the tensor field for two reasons. Firstly, since our proposed method works on a raster image, it is imperative that the image must be of high resolution. While this is guaranteed by chart images procedurally/programmatically generated, the images readily available from various sources or scanned from the textbook could be of lower quality. Conventional image processing includes compression, formatting, smoothening, etc. One of the undesirable outcomes of these processing methods is aliasing, which affects our raster-based methods. However, we can preprocess the chart image by performing antialiasing on the component boundary. Secondly, we require only the chart canvas from the chart image for data extraction. Thus, we remove all other chart components, namely, axes, gridlines, legend, and text for data extraction. Overall, we preprocess for antialiasing and chart canvas extraction. 

We perform tasks such as grid removal and adding a border to bars to preprocess the chart images, implemented using image binarization followed by morphological operations (Algorithm~\ref{alg:canvas}). Morphological operations are a set of image processing operations that exploit the property of shape. They are used routinely for extracting connected components from images that are useful in representation and description of region shape. We first check if the chart objects are filled or not, by binarizing and checking the percentage of filled pixels. If the chart objects are not filled, we perform object-fill, as our proposed tensor fields are effective for data extraction only in charts with filled chart objects. The morphological operations for foreground extraction involve erosion for white noise removal, dilation for expanding shrunk objects, and appropriate distance transformations. We perform the morphological opening of the binarized image and dilation for background extraction. The foreground is used for extracting connected components, which is followed by the watershed algorithm used for segmentation. The segmented image is the subset of chart canvas, \ie the chart-object-clusters.

These morphological operations remove formatted borders, if present, of all chart-object-clusters. However, the differences in color between component interior, component exterior, and component boundary manifest as strong degenerate points in corners or junctions. Such degenerate points help in improving the accuracy of the data extraction. Hence, in our image preprocessing procedure, we further perform contouring to identify component boundary and add a border of predetermined pixel-width. This ensures borders for all chart-object-clusters, uniformly across input chart images. We have found 2-pixel-wide borders to be effective. 

\begin{algorithm}[htbp]
  \caption{Chart canvas extraction}
  \label{alg:canvas}
  \DontPrintSemicolon
  
\KwInput{Chart image $C_i$}
\KwOutput{Chart canvas $C_c$}
$C_b$ $\leftarrow$ binarize($C_i$) \;
$n_w$ $\leftarrow$ count-white-pixels($C_b$) \;
$n_b$ $\leftarrow$ count-black-pixels($C_b$) \;
\If {$n_w$ < $0.2*(n_w+n_b)$}
{
  $C_i$ $\leftarrow$ object-fill($C_i$) \;
  $C_b$ $\leftarrow$ binarize($C_i$) \;
}
 \tcc{Morphological operations for foreground} \;
$C_{fg}$ $\leftarrow$ morphology-foreground-extraction($C_b$)\;
$C_b$ $\leftarrow$ morphology-opening($C_b$) \tcp*{Morphological opening operation}\;
$C_{bg}$ $\leftarrow$ morphology-dilation($C_b$)\tcp*{Morphological dilation for background}\;
components, component-labels $\leftarrow$ find-connected-components($C_{fg}$)\;
\For {pixel i in $C_{bg}$}
{
  component-labels(i) := 0 \tcp*{Setting component label for background}\;
 }
$C_\textrm{segmented}$ $\leftarrow$ morphology-watershed($C_i$, component-labels) \tcp*{Object extraction}\;
object-edges $\leftarrow$ Canny-edge-detection($C_\textrm{segmented}$) \tcp*{Edge detection}\;
contours := contouring(object-edges, pixel-width=2) \tcp*{Contour extraction}\;
\For {pixel i in contours}
     {
       color($C_\textrm{segmented}$(i)) := black \tcp*{Adding 2-pixel width border to objects}\;
       }
\KwRet $C_\textrm{segmented}$\;
\end{algorithm}

\mysubsubheading{Error Analysis of Extracted Data:}
We qualitatively compare the visualizations of the chart of the same type plotted using the extracted data and the original chart for each experiment. For quantitative evaluation, we must first determine the performance of chart reading, which is tied to the accuracy with which quantitative information can be ``decoded'' from the chart specifiers, such as geometry~\cite{carswell1990perceptual}. In our case, we observe that our proposed model is effective only if the error in data extraction is minimal. Hence, we determine the error between our extracted data and the ground truth. However, for quantitative comparisons, we run into the issue of our data being extracted in image space, owing to the loss of context of the chart provided by the text, axes, and legend. Overall, we compare distributions of extracted data with the ground truth by using appropriate distance measures.

Earth mover's distance, $\emd$, is a measure of cross-bin distances between histograms of distributions, which uses ground distance measures~\cite{rubner1997earth}. We compute $\emd$ between the univariate distributions for the extracted data and original data in the case of bar charts, $\emdbc$. In the case of scatter plots, we compute the $\emd$ between the 2D point clouds of the extracted data and original data, $\emdsp$.

\section{Experiments \& Results}\label{sec:expt}
\begin{figure}[htbp]
\centering
\fbox{\includegraphics[width=\columnwidth]{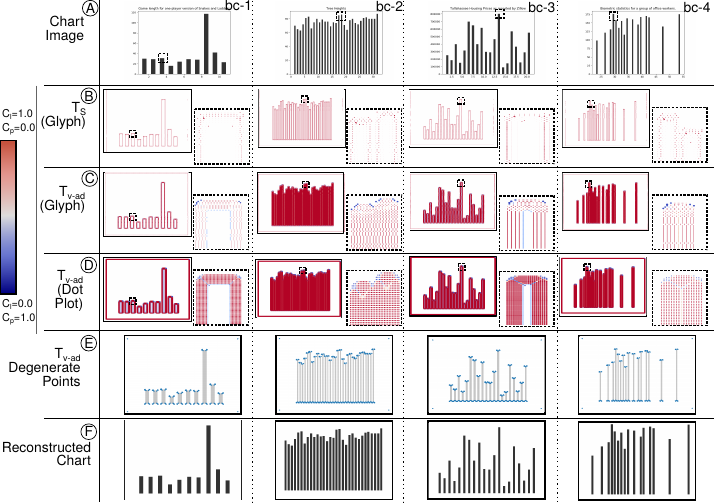}}
\caption{\label{fig:dst-bc} Our proposed data extraction method from programmatically generated raster images of bar charts from readily available table data (\dst), using saliency map of structure tensor $\st$ and tensor voting $\tvad$, to determine degenerate points. }
\end{figure}

\begin{figure}[htbp]
\centering
\fbox{\includegraphics[width=\columnwidth]{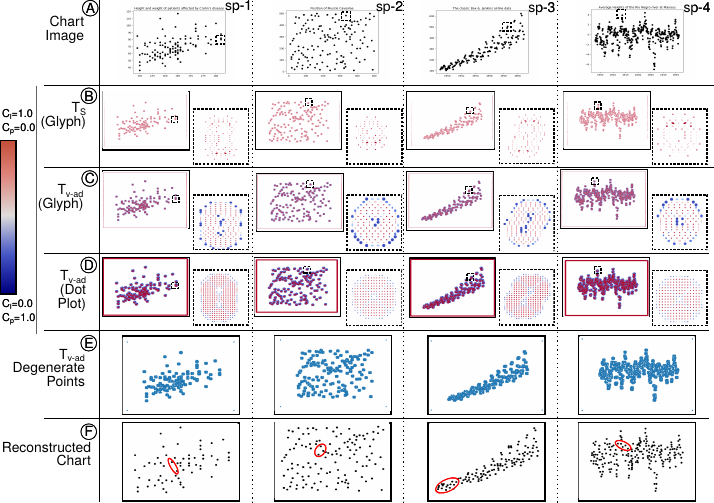}}
\caption{\label{fig:dst-sp} Our proposed data extraction method from programmatically generated raster images of scatter plots from readily available table data (\dst), using saliency map of structure tensor $\st$ and tensor voting $\tvad$, to determine degenerate points. The reconstruction errors are marked in red in row F.}
\end{figure}

\begin{figure}[htbp]
\centering
\fbox{\includegraphics[width=\columnwidth]{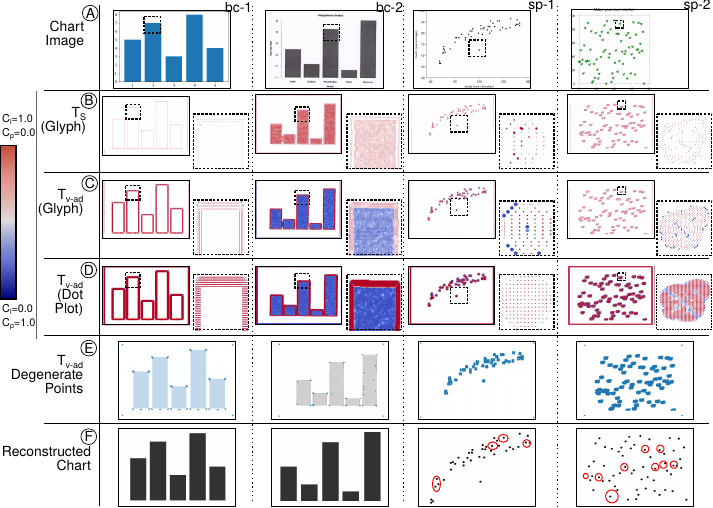}}
\caption{\label{fig:dsi-bc-sp} Our proposed data extraction method from raster images of bar charts and scatter plots from readily available image data (\dsi), using saliency map of structure tensor $\st$ and tensor voting $\tvad$, to determine degenerate points. The reconstruction errors are marked in red in row F.}
\end{figure}

\begin{figure}[htbp]
\centering
\fbox{\includegraphics[width=\columnwidth]{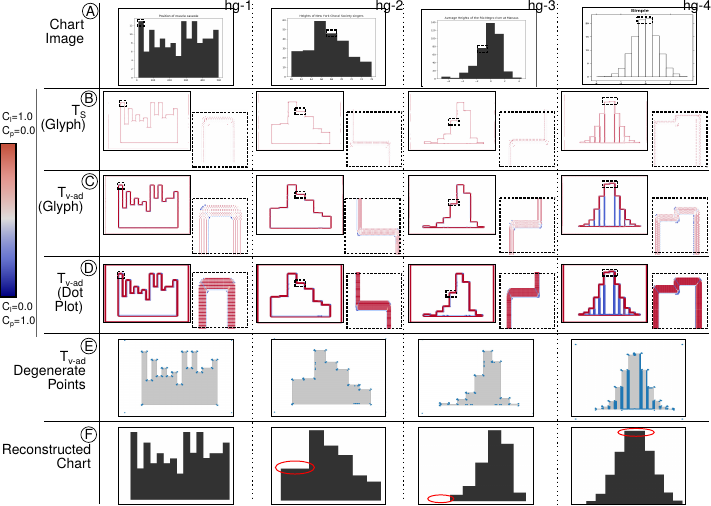}}
\caption{\label{fig:dst-dsi-hg} Our proposed data extraction method from raster images of histograms, using saliency map of structure tensor $\st$ and tensor voting $\tvad$, to determine degenerate points, of (a) readily available table data (\dst) \{hg-1, hg-2, hg-3\}, and (b) chart images (\dsi), \{hg-4\}. The reconstruction errors are marked in red in row F.}
\end{figure}

For experiments, we have performed data extraction using tensor fields on bar charts and scatter plots, as well as on histograms, as a special case of bar charts. We have used three sets of data for our experiments. The dataset descriptions are available at the project GitHub webpage\footnote{\url{https://github.com/GVCL/Tensor-field-framework-for-chart-analysis}}. The dataset \dst~contains multivariate table datasets that are publicly available, and \dsi~contains chart images that are publicly available. We have programmatically generated charts for \dst~using Python library, \texttt{matplotlib.pyplot}~\cite{hunter2007matplotlib}, and stored them in .png image format. We have specifically used this library, as it generates high-resolution images, compared to a plotting tool, such as Microsoft\textregistered Excel\textregistered. We have reported the dataset sources in the Acknowledgements.

For all chart images of test datasets, we have constructed the tensor fields and reconstructed data. For bar charts, we have tested with datasets with a large number of bars (\ie thinner bars), non-uniformly placed bars, smaller set of bars, and bar charts with large variation in the bar heights. For scatter plots, we have tested for positive, negative, and zero correlation data, with a large number of scatter points, with overlapping scatter points. As a special case of simple bar charts, we have used a dataset for histograms. In the case of histograms, we have tested with datasets with variations in the number of bins, with close-to-zero frequencies in some of the histogram bins, with several large variations in frequencies (\ie several peaks and valleys in the histogram), and with the close-to-normal distribution. Since we have the ground truth for dataset \dst, we have performed an error analysis of the same. For the error analysis of histograms, we have determined the Earth mover's distance between the extracted data and the frequency table of the original data, $\emdhg$.

\begin{table}[htbp]
\centering
\begin{tabular}{|c|c||c|c||c|c|}
\hline
\multicolumn{2}{|c||}{Bar Chart (\dst) }
& \multicolumn{2}{c||}{Scatter Plot (\dst)} 
& \multicolumn{2}{c|}{Histogram (\dst)}
\\\hline
(Figure~\ref{fig:dst-bc}) & $\emd$ 
& (Figure~\ref{fig:dst-sp}) & $\emd$
 &(Figure~\ref{fig:dst-dsi-hg}) & $\emd$ 
\\\hline
bc-1 & 3.5e-4
& sp-1 & 5.4e-2
& hg-1 & 1.9e-2
\\\hline

bc-2 & 4.4e-3
& sp-2 & 1.1e-2
& hg-2 & 6.3e-3
\\\hline

bc-3 & 2.6e-3
& sp-3 & 5.5e-2
& hg-3 & 3.9e-2
\\\hline

bc-4 & 3.0e-3
& sp-4 & 2.7e-2
& \multicolumn{2}{c}{}
\\\cline{1-4}

\end{tabular}
\caption{\label{tab:error}
Reconstruction error using Earth Mover's Distance of distributions of normalized values in data tables from source and that reconstructed from chart images.}
\end{table}

Our results are provided in Figures~\ref{fig:dst-bc},~\ref{fig:dst-sp},~\ref{fig:dsi-bc-sp},~\ref{fig:dst-dsi-hg}, and Table~\ref{tab:error}. We describe aspects of our results related to the tensor field model, data extraction, and error analysis here.

\mysubsubheading{Tensor Field Model:} Starting from the pixel-wise gradient tensors in the chart images, the tensor voting field after anisotropic diffusion gives stronger degenerate points (\ie higher $\cp$ values) than the structure tensor. This is uniformly observable in all the experiments (Figures~\ref{fig:dst-bc}--\ref{fig:dst-dsi-hg}). Hence, we choose to use $\tvad$ further for data extraction.

The glyph-based tensor field visualization of the subsampled grid  (Figures~\ref{fig:dst-bc}--\ref{fig:dst-dsi-hg}, row D) is effective for locating strong degenerate points. The 
dot plots  (Figures~\ref{fig:dst-bc}--\ref{fig:dst-dsi-hg}, rows B-C) of the tensor fields using color map based on  the saliency map are effective for overall tensor field visualization. 

The thickness of the bar influences tensor field modeling. In thicker bars (Figure~\ref{fig:dst-bc}), we observe that in the component interior of the bar object has a region of zero-tensor near the centroid. We refer to this as the region of ``homogeneity'', where tensors have zero-gradient value. The homogeneous region near the center of the bar grows with its thickness.

\mysubsubheading{Data Extraction:} Our results for bar charts (Figures~\ref{fig:blockdiagram},~\ref{fig:dst-bc} and,~\ref{fig:dsi-bc-sp}) show that our method can extract data for a variety of charts stored in both image format and those programmatically plotted from available data tables. Using morphological operations has been effective in removing aliasing effects in low-quality chart images (Figure~\ref{fig:blockdiagram} and \texttt{bc-2} in Figure~\ref{fig:dsi-bc-sp}). In addition to preprocessing, postprocessing by filtering out the weak degenerate points helps in denoising. We have used filtering threshold $\tau_{wd}=0.003$ for histograms, different from bar charts.

\mysubsubheading{Error Analysis:} In the case of scatter plots, our proposed data extraction method suffers from omission (type-2) errors, \ie false negatives (Figures~\ref{fig:dst-sp} and~\ref{fig:dsi-bc-sp}, row F). This happens when scatter points are clustered together or when the mark of the scatter point intersects the axes (Figure~\ref{fig:dsi-bc-sp}, \texttt{sp-2}). Our model requires clustering of degenerate points to localize a single scatter point, and the second level of clustering to separate clusters of scatter points. DBSCAN parameters have to be modified for each run depending on the pixel-distances between the scatter points.

We compare our result of data extraction in scatter plots with that of Scatteract~\cite{cliche2017scatteract} (Figure~\ref{fig:dsi-bc-sp}, \texttt{sp-2}), which had several false positives (type-1 errors). The type-1 errors have been attributed to each ``clover'' mark being detected as three different scatter points. In our case, we have detected the ``clover'' marks as centroids of clusters. Thus, we avoided false positives but not omission errors.

As a special case of bar charts, histograms perform well with our proposed data extraction method (Figure~\ref{fig:dst-dsi-hg}). We observe errors in the case of histograms with two adjacent bars with similar heights  (Figure~\ref{fig:dst-dsi-hg}, \texttt{hg-2} and, \texttt{hg-4}), as well as with bars that are close to the x-axis (Figure~\ref{fig:dst-dsi-hg}, \texttt{hg-3}).

The error analysis using $\emd$ shows higher errors in the case of scatter plots than the bar charts, or histograms, owing to the type-2 errors (Table~\ref{tab:error}). In the case of histograms, the histograms of tailed distributions containing $\sim$0-frequency bins at the tails cause larger errors (Figure~\ref{fig:dst-dsi-hg}, \texttt{hg-3}).

\mysubheading{Discussion:}
We explain the different factors that influence our tensor field construction to evaluate the robustness of such a field. We also discuss the alternative school of thought, the Gestalt theory of perceptual organization for chart interpretation.

\begin{figure}[thbp]
\centering
\includegraphics[width=\columnwidth]{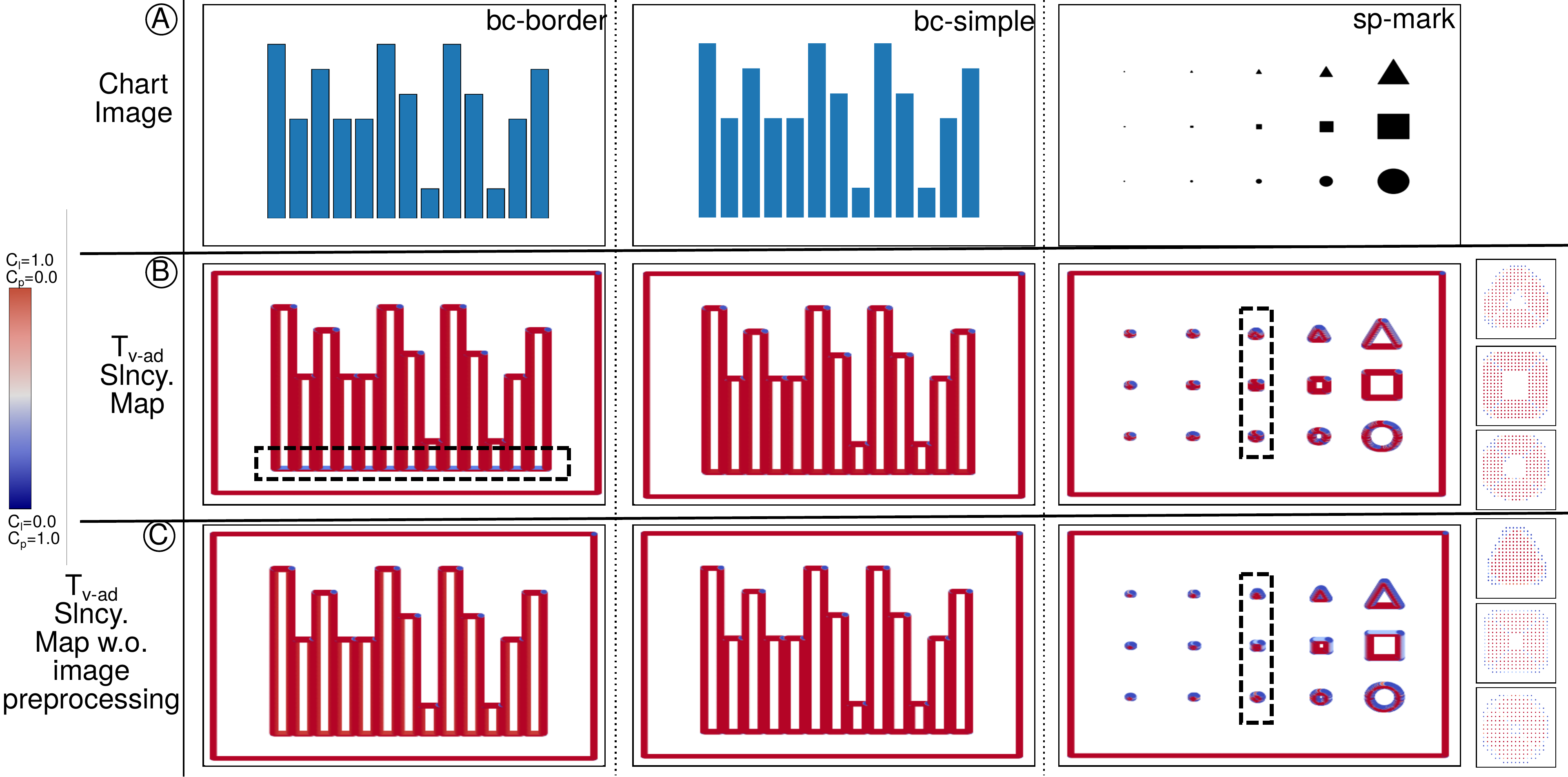}
\caption{\label{fig:discussion} The differences in the saliency (Slncy.) map of tensor voting with anisotropic diffusion $\tvad$ with and without image preprocessing on chart objects: (left, middle) bars in bar chart with and without formatted border, (right) marks in scatter plot.}
\end{figure}

\mysubsubheading{Geometry:} Our results in bar charts (Figure~\ref{fig:dst-bc}) show the effect of bar width in the tensor field computation. In a similar vein, we tested the influence of the size of the mark of scatter points on the tensor field (Figure~\ref{fig:discussion}, \texttt{sp-mark}). We observe and confirm that the homogeneous regions grow larger in the center/centroid of chart-object-clusters with the size of the connected component(s). The position of the chart objects, and hence, that of the chart-object-clusters, is ideally the centroid of the homogeneous region. However, the homogeneous region itself does not carry any information, and the degenerate points occur in the non-homogeneous region surrounding the homogeneous region in the chart-object-cluster. Thus, the position of the chart-object-cluster is approximated as the centroid of the clusters of the degenerate points in the non-homogeneous region, as done in our proposed method (Algorithm~\ref{alg:workflow}). Overall, the chart object size does not affect our data extraction results.

For scatter plots, we observe that all shapes of scatter points have degenerate points in their pixelated format (Figure~\ref{fig:discussion}, \texttt{sp-mark}). While shapes with inherent corners reinforce degenerate point clusters, we also observe that the distribution of these clusters is not uniform in such cases. Hence, this introduces minor errors in the position of the chart object. We have already discussed the type-2 errors in our method owing to overlapping or clustered scatter points. It may be noted that not all type-2 errors lead to a change in correlation, which is an of-studied statistical measure using scatter plots. We expect similar type-2 errors in bar charts where thin bars are closely placed. The degenerate-point-clustering algorithm needs to be evaluated and modified to handle the chart object localization and separation automatically.

Our method currently works for charts with separable geometric objects, such as bars, scatter plots,  and with corners. In such charts, the tensor fields produce degenerate points whose location corresponds directly to the value of the data to be extracted. Further study needs to be undertaken on the role of degenerate points in the case of charts where there is non-linear mapping of the geometric object properties to data, e.g. sector area in the case of pie charts.

\mysubsubheading{Color:} Since the tensor field is computed on raster images which have a color attribute, the color model and the color difference functions used for generating the gradient tensor influence the local geometric descriptors. The CIELab color model has been recommended to be the model for computing tensor voting for color image denoising~\cite{moreno2011edge}. In our work, we have used either the grayscale or 1-color palette, which requires only one of the 3 channels in the RGB model for computing gradients. The influence of color and in a related way, texture are yet to be studied in depth in chart image processing for data extraction.

\mysubsubheading{Image Preprocessing:} We have observed that the morphological operations performed to address the influence of aliasing and other image formatting help in reducing noise in the data (Figure~\ref{fig:dsi-bc-sp}, \texttt{bc-1} and, \texttt{bc-2}). Our image preprocessing procedure reintroduces border for all chart-object-clusters, owing to which we expect generalized behavior of tensor fields and its degenerate points in the bar and scatter point objects. However, we observe that in the case of bar charts with formatted borders, our new borders introduce degenerate points in the base of the bar objects (Figure~\ref{fig:discussion}, \texttt{bc-border}, row B). This experiment was conducted on programmatically generated bar charts (Figure~\ref{fig:discussion}, \texttt{bc-border}, \texttt{bc-simple}) to remove influences of other image processing or compression artefacts.

Our method is designed for raster images using gradient tensor field processing for edge detection. The data extracted from these images is interpreted using the shapes formed by these edges, and hence, the data is WYSIWYG. That also implies that our method works only for discernible images. Our method, thus, fails in the case of images with noisy content and/or of low resolution, which cause the images to be indiscernible.

\mysubsubheading{Machine Learning Models:} While image processing needed for computer vision applications is done using robust machine learning models, chart image processing is an understudied area. Most models work for specific chart type(s)~\cite{choi2019visualizing, cliche2017scatteract}, and unified models for larger classes of chart types are still a gap. This area can be better served using appropriate machine modeling constructs. The characterization of the chart images and study of feature extraction from such data enables the construction of appropriate models that handle sparsity and geometric characterization in images appropriately. In terms of sparsity, the machine learning models must also account for the data-ink ratio in chart images. The images used for our experiments, the data-ink ratio has been less than 10\%, which is considerably low.

\mysubsubheading{Emergent Features:} There are two different kinds of perceptual information, namely local and emergent features. Local features are those computed in local proximity in an image, and emergent features are non-decomposable and higher-order ones, which provide an overview/global understanding of the image. For instance, in a scatter plot, a single scatter point is a local feature, whereas the grouping of the scatter points implies the type of correlation, and hence is an emergent feature. These features have their respective theories governing how they contribute to the perceptual understanding of images. The Gestalt theory of perceptual organization states that emergent features enable in the perceptual understanding of the image, and the local theory emphasizes on the local features, instead. The Gestalt theory has been experimentally found to be a better model for perceptual understanding than the local theory for a pair of dots~\cite{hawkins2016can}. In cognitive science, this theory is rephrased as the Principle of Perceptual Organization~\cite{hegarty2011cognitive}, which says, 
\begin{quote}
  ``Ensure that groupings based on Gestalt principles, and emergent features more generally, are compatible with the tasks to be carried out with a display.''
\end{quote}

In our work, we have focused on the local theory. Our overarching goal is to model the Gestalt theory. We chose tensor voting for generating the tensor field for its extensibility to the Gestalt model, as tensor voting takes into consideration Gestalt principles of the perceptual organization~\cite{medioni2000tensor}. 

\section{Conclusions}\label{sec:conclusions}
Automating chart interpretation has been a long-standing challenge owing to the vast design space of charts and their raster images. We have demonstrated automating data extraction from chart images using a computational model that exploits the local structure in the raster images. The novelty of our work lies in identifying tensor voting after anisotropic diffusion, $\tvad$, as an effective local geometric descriptor for data extraction. We have used the positive semidefinite second-order tensor fields of the local geometric descriptors in the chart canvas and extracted the tensor topological features, namely degenerate points, for data extraction. We have shown how patterns of clustering of degenerate points correspond to the chart data, enabling data extraction. We focus on bar charts and scatter plots where the proximity of local features defines the charts' geometric objects. There are limitations to our work as the tensor field is dependent on the user-defined image characteristics. These include image resolution and styling features of the plots, e.g., glyph shapes and sizes, bar width, and borders.

Our current model achieves Levels-A1 and A2 in Kimura's six-level scheme of statistical ability~\cite{aoyama2003graph} by performing data extraction with considerable accuracy. The future scope of our work is in automated clustering and cluster-classification of degenerate points and extending the usefulness of the tensor voting for Gestalt theory of perceptual organization~\cite{medioni2000tensor} in charts. This extension will enable the computational model to achieve the Level-A in Kimura's six-level scheme of statistical ability~\cite{aoyama2003graph}.

\section*{Acknowledgment}
The authors acknowledge the financial support from the MINRO (Machine Intelligence and Robotics) grant from the Government of Karnataka, for this work. The authors are grateful to all members of GVCL and IIIT Bangalore for supporting this work. Several people have helped shape this work through feedback and discussions: anonymous reviewers, T. K. Srikanth, IIITB; Sindhu Mathai of Azim Premji University; Vidhya Y. and Supriya Dey of Vision Empower; Neha Trivedi, XRCVC; Vani, Pushpaja, Kalyani, and Anjana of Braille Resource Center, Matruchayya, that has shaped this work.

The datasets used in \dst are from CSV data repository \url{https://people.sc.fsu.edu/~jburkardt/data/csv/csv.html} for bar charts, and from R datasets archive~\cite{bundock2012rdatasets} for histograms and scatter plots. The datasets used in \dsi are from online documentation of package BoutrousLab.plotting.general~\cite{png2019guide}, except for one histogram from a course project in ``Computer Vision'' course by Roberts~\cite{lozman1997pre}. The bar chart available at \url{http://www.datasciencemadesimple.com/r-bar-chart/} has been printed and scanned to get the chart image \texttt{bc-2} in \dsi.

\bibliographystyle{unsrt}
\bibliography{papers_consolidated}

\end{document}